\documentclass[conference]{IEEEtran}
\IEEEoverridecommandlockouts
\usepackage{cite}
\usepackage{amsmath,amssymb,amsfonts}
\usepackage{algorithmic}
\usepackage{graphicx}
\usepackage{multirow}
\usepackage{textcomp}
\usepackage{booktabs} 
\usepackage{xcolor}
\def\BibTeX{{\rm B\kern-.05em{\sc i\kern-.025em b}\kern-.08em
    T\kern-.1667em\lower.7ex\hbox{E}\kern-.125emX}}
\begin{document}

\title{Small Object Detection in Complex Backgrounds with Multi-Scale Attention and Global Relation Modeling\\
}

\author{\IEEEauthorblockN{1\textsuperscript{st} Wenguang Tao}
\IEEEauthorblockA{\textit{Unmanned System Research Institute} \\
\textit{Northwestern Polytechnical University}\\
Xi'an, China \\
wenguangtao2022@mail.nwpu.edu.cn}
~\\
\and
\IEEEauthorblockN{2\textsuperscript{nd} Xiaotian Wang*}
\IEEEauthorblockA{\textit{Unmanned System Research Institute} \\
\textit{Northwestern Polytechnical University}\\
Xi'an, China \\
18710993786@163.com}
*Corresponding author
~\\
\and
\IEEEauthorblockN{3\textsuperscript{rd} Tian Yan}
\IEEEauthorblockA{\textit{Unmanned System Research Institute} \\
\textit{Northwestern Polytechnical University}\\
Xi'an, China \\
tianyan@nwpu.edu.cn}
~\\
\and
\IEEEauthorblockN{4\textsuperscript{th} Yi Wang}
\IEEEauthorblockA{\textit{Department of Electrical and Electronic Engineering} \\
\textit{The Hong Kong Polytechnic University}\\
Hong Kong, China \\
yi-eie.wang@polyu.edu.hk}

\and
\IEEEauthorblockN{5\textsuperscript{th} Jie Yan}
\IEEEauthorblockA{\textit{Unmanned System Research Institute} \\
\textit{Northwestern Polytechnical University}\\
Xi'an, China \\
jyan@nwpu.edu.cn}

\and

}

\maketitle

\begin{abstract}
Small object detection under complex backgrounds remains a challenging task due to severe feature degradation, weak semantic representation, and inaccurate localization caused by downsampling operations and background interference. Existing detection frameworks are mainly designed for general objects and often fail to explicitly address the unique characteristics of small objects, such as limited structural cues and strong sensitivity to localization errors. In this paper, we propose a multi-level feature enhancement and global relation modeling framework tailored for small object detection. Specifically, a Residual Haar Wavelet Downsampling module is introduced to preserve fine-grained structural details by jointly exploiting spatial-domain convolutional features and frequency-domain representations. To enhance global semantic awareness and suppress background noise, a Global Relation Modeling module is employed to capture long-range dependencies at high-level feature stages. Furthermore, a Cross-Scale Hybrid Attention module is designed to establish sparse and aligned interactions across multi-scale features, enabling effective fusion of high-resolution details and high-level semantic information with reduced computational overhead. Finally, a Center-Assisted Loss is incorporated to stabilize training and improve localization accuracy for small objects. Extensive experiments conducted on the large-scale RGBT-Tiny benchmark demonstrate that the proposed method consistently outperforms existing state-of-the-art detectors under both IoU-based and scale-adaptive evaluation metrics. These results validate the effectiveness and robustness of the proposed framework for small object detection in complex environments.
\end{abstract}


\section{Introduction}
Small object detection remains a long-standing and challenging problem in computer vision, particularly under complex backgrounds and adverse imaging conditions \cite{1}. Due to their limited spatial extent, weak visual contrast, and susceptibility to background noise, small objects often suffer from severe feature degradation during convolutional downsampling, leading to ambiguous semantic representations and unstable localization \cite{2}. These challenges are further amplified in real-world scenarios such as aerial surveillance, where objects frequently occupy only a few pixels and appear under diverse illumination conditions \cite{3}.

Recent advances in object detection have largely focused on improving detection performance through deeper backbones, multi-scale feature pyramids, and attention-based feature fusion \cite{4}. However, most existing detectors are primarily designed for general object detection and rely on generic backbone architectures, which inadequately account for the unique characteristics of small objects \cite{5}. In particular, conventional downsampling operations tend to discard fine-grained structural details, while standard feature aggregation strategies often ignore the nonlinear spatial correspondence across different scales. As a result, small object features are easily overwhelmed by background responses, leading to degraded detection accuracy.

To alleviate these challenges, numerous studies have explored specialized solutions for small object detection. Hou et al. \cite{6} combined handcrafted feature representations with convolutional neural networks to establish a mapping between feature maps and the likelihood of small targets in images, enabling effective segmentation of true weak targets. Wang et al. \cite{wang2021self} proposed a multi-scale density map regressor \cite{wang2019object} and further developed Crowd-SDNet, which enables the estimation of object center locations and sizes for densely packed objects using only point-level annotations. Xu et al. \cite{7} designed novel frequency–spatial convolutions and employed Haar wavelet transforms to decompose input features, thereby emphasizing discriminative information of small objects. Tang et al. \cite{8} developed a dual-attention Transformer-based model that integrates infrared and low-light visible images, achieving improved performance during sampling and pooling operations. Yuan et al. \cite{9} enriched small object representations by incorporating prior knowledge with data-driven semantic interactions, effectively suppressing background noise. In addition, Ying et al. \cite{3} proposed a scale-adaptive fitness (SAFit) metric, which demonstrates high robustness in small object regression tasks. While these approaches demonstrate promising improvements, they are often applied in isolation. Moreover, existing attention mechanisms typically incur high computational costs or fail to explicitly address cross-scale feature alignment, limiting their effectiveness in resource-constrained detection pipelines.

In this paper, we propose a multi-level feature enhancement and global relation modeling framework tailored for small object detection in complex environments. First, a Residual Haar Wavelet Downsampling module is introduced to jointly exploit spatial convolutional features and frequency-domain high- and low-frequency components, effectively alleviating detail loss caused by downsampling. Second, a Global Relation Modeling module is employed at the high-level feature stage to capture long-range dependencies and suppress background interference, providing stable global semantic priors for downstream tasks. Third, a Cross-Scale Hybrid Attention module is designed to establish sparse yet aligned interactions across multi-scale feature maps, enabling effective fusion of high-resolution details and high-level semantics with reduced computational overhead. Finally, a Center-Assisted Loss is incorporated into the regression branch to enhance training stability and improve localization accuracy for small objects, particularly when IoU-based supervision becomes ineffective.

Extensive experiments conducted on the large-scale RGBT-Tiny benchmark demonstrate the effectiveness and robustness of the proposed framework. The results demonstrate that our method achieves state-of-the-art performance. The main contributions of this work can be summarized as follows:
\begin{itemize}
	\item We propose a unified small object detection framework that jointly enhances structural details, global semantic reasoning, and cross-scale feature alignment. Our method achieves state-of-the-art performance on the RGBT-Tiny dataset.
	\item The residual wavelet-based downsampling strategy is introduced to preserve fine-grained information during early feature extraction.
	\item The global relation modeling module is proposed to aggregate long-range global semantic information.
	\item The cross-scale hybrid attention module is designed to achieve efficient and accurate multi-scale feature fusion.
\end{itemize}

\section{Methodology}
\begin{figure*}
	\centering
	\includegraphics[width=\textwidth]{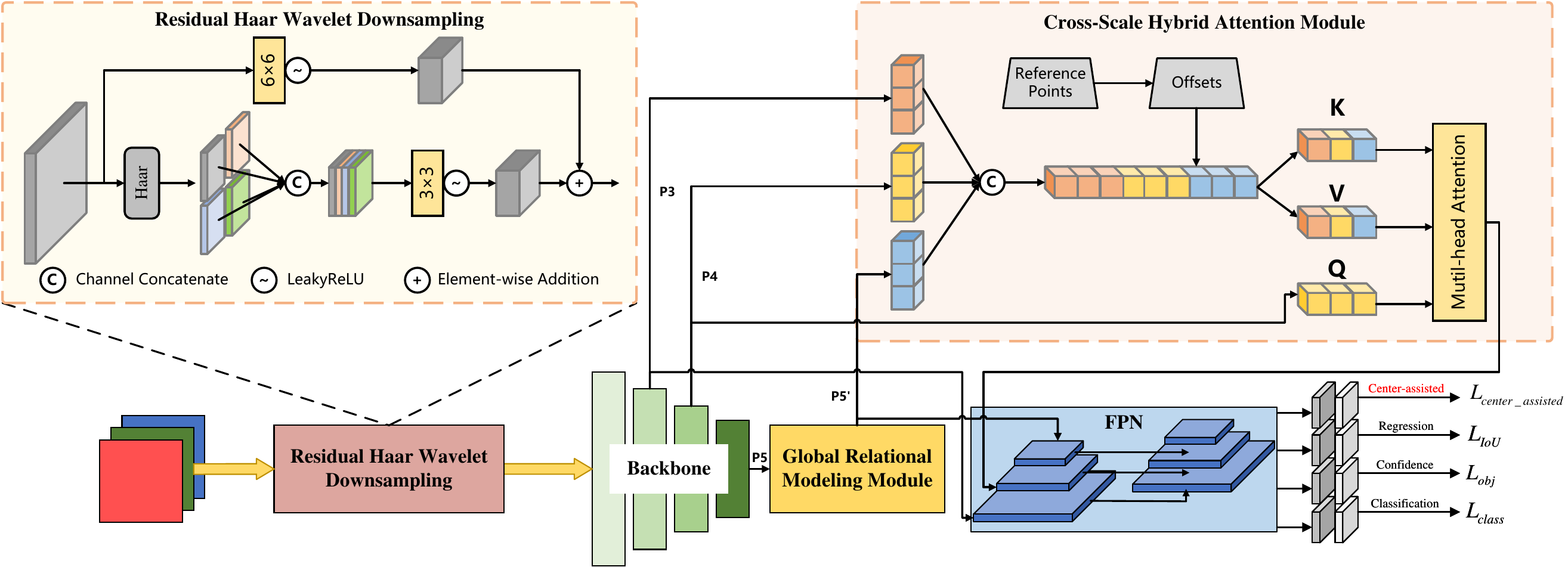}
	\caption{Overall Framework Diagram. The input image is fed into the backbone network for feature extraction after passing through the residual Haar wavelet downsampling module (RHWD). Subsequently, the P3, P4, and P5 features are respectively input into the global relation modeling module (GRM) and cross-scale hybrid attention module (CSHA) for feature enhancement, and the detection results are finally obtained via the feature pyramid network (FPN).}
	\label{fig1}
\end{figure*}

To address the challenges of insufficient structural information, ambiguous semantic representation, and misalignment across multi-scale features in complex backgrounds, this paper proposes a multi-level feature enhancement and global relationship modeling framework for small object detection. By incorporating global semantic modeling, multi-scale feature aggregation, and regression optimization strategies, the proposed framework effectively improves the detection performance of small objects. The overall architecture is illustrated in Fig.~\ref{fig1}.

During the feature extraction stage, a residual Haar wavelet downsampling module is introduced to alleviate the loss of fine-grained details of small objects caused by downsampling, by jointly exploiting spatial convolutional features and frequency-domain low- and high-frequency information. At the high-level semantic stage, a global relation modeling module is employed to capture long-range dependencies in deep features, providing stable global semantic priors for the network. Subsequently, a cross-scale hybrid attention module establishes sparse and aligned dynamic interactions across multi-scale features, effectively integrating high-resolution details with high-level semantic information. Finally, a center-assisted loss is incorporated as an auxiliary constraint for bounding box regression to enhance training stability and improve localization accuracy for small objects.

\subsection{Residual Haar Wavelet Downsampling}

Small objects suffer from limited discriminative information, weak contrast against background clutter, and severe noise interference. In general, richer structural cues lead to better detection performance. However, most existing small object detection networks rely on generic backbone architectures for feature extraction, largely overlooking the unique characteristics of small objects. To address this limitation, we propose a residual Haar wavelet downsampling module (RHWD).

RHWD takes the raw input image and processes it through two parallel branches. The global branch employs large-receptive-field convolutions to capture abstract and generalized object representations, while the local branch applies the Haar wavelet transform to extract frequency-domain low- and high-frequency components. The resulting features from both branches are then fused via element-wise addition to achieve complementary feature integration, yielding a more expressive hybrid representation.

Specifically, the input image $I\in{\mathbb{R}^{H \times W \times 3}}$ is processed by two parallel branches, namely a global branch and a local branch, to jointly model long-range contextual dependencies and fine-grained edge details:
\begin{equation}
\left\{ \begin{gathered}
	{F_{Global}} = \delta ({f_{BN}}(Con{v_{6 \times 6}}(I))) \hfill \\
	{A_c},{H_c},{V_c},{D_c} = {\phi _{Haar}}(I) \hfill \\
	{F_{Local}} = Cat({A_c},{H_c},{V_c},{D_c}) \hfill \\ 
\end{gathered}  \right.
\end{equation}
where $\mathrm{Conv}_{6\times6}(\cdot)$ denotes a $6\times6$ convolution operation with a stride of 2 for downsampling, ${f_{BN}}$ represents batch normalization, and $\delta(\cdot)$ is the SiLU activation function. ${\phi _{Haar}}(\cdot)$ denotes the Haar wavelet transform, ${A_c}$,${H_c}$,${V_c}$,${D_c}$ are one low-frequency approximation component and three high-frequency detail components, $\mathrm{Cat}(\cdot)$ indicates channel-wise concatenation.

Subsequently, the concatenated frequency-domain features are projected to the same channel dimension and fused with the global semantic features through a residual manner:
\begin{equation}
{I_{fuse}} = \delta (Con{v_{3 \times 3}}({F_{Local}})) + {F_{Global}}
\end{equation}
where $\mathrm{Conv}_{3\times3}(\cdot)$ denotes a $3\times3$ convolution for channel expansion and feature alignment.

By jointly exploiting spatial-domain convolutional features and frequency-domain representations, RHWD effectively preserves fine-grained details during downsampling. This complementary fusion enables consistent semantic alignment between spatial and frequency features, thereby enhancing the representation capability for small objects.

\subsection{Global Relation Modeling Module}

\begin{figure}[htbp]
	\centering
	\includegraphics[width=\columnwidth]{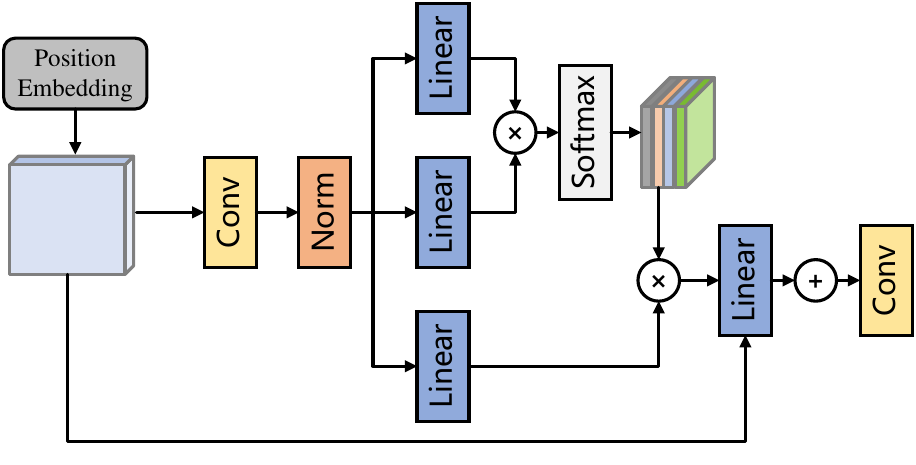}
	\caption{Global relation modeling module (GRM).}
	\label{fig2}
\end{figure}

Conventional convolutional backbones lack the ability to model long-range dependencies and are therefore susceptible to background noise when detecting small objects. The deepest layers of the network typically encode abstract semantic representations that are scarce in shallow features. To better aggregate global dependencies across multiple scales, we introduce a global relation modeling module (GRM) at the end of the backbone network. This module suppresses background feature responses and provides global semantic priors for downstream tasks, enabling the model to focus on candidate regions containing small objects and further enhancing fine-grained and discriminative feature representations, as illustrated in Fig.~\ref{fig2}.

Specifically, GRM consists of layer normalization, multi-head self-attention, and residual connections. First, the P5 feature map is flattened into a sequence ${X_{P5}} \in {\mathbb{R}^{B \times (H \times W) \times C}}$, and combined with a learnable positional embedding before being fed into a normalization layer:
\begin{equation}
	{F_{embed}} = {f_{BN}}({X_{P5}} + {E_{pos}})
\end{equation}
where ${E_{pos}}$ denotes the learnable positional embedding. The embedded features ${F_{embed}}$ are then linearly projected to obtain the Query, Key, and Value matrices. For the i-th attention head, the computation is defined as:
\begin{equation}
\left\{ \begin{gathered}
	{Q_i},{K_i},{V_i} = {F_{embed}} \cdot ({W_Q},{W_K},{W_V}) \hfill \\
	{\text{Attention(}}{Q_i},{K_i},{V_i}{\text{)}} = {\text{Softmax}}\,\left( {\frac{{{Q_i} \cdot {K_i}}}{{\sqrt d }}} \right) \cdot {V_i} \hfill \\ 
\end{gathered}  \right.
\end{equation}
where $\sqrt d$ is the scaling factor. The outputs of all attention heads are concatenated and projected through a linear layer to enhance multi-scale feature interactions:
\begin{equation}
	{Z_i} = {\text{Concat}}\left[ {{\text{Attentio}}{{\text{n}}_1}, \ldots ,{\text{Attentio}}{{\text{n}}_i}} \right] \cdot {W_O}
\end{equation}

Finally, a residual connection is employed to add the attention-enhanced features back to the original features, alleviating gradient vanishing and preserving semantic consistency. GRM effectively aggregates image-level contextual information and propagates global semantic cues to lower-level features through the FPN in a top-down manner. Moreover, it provides essential semantic support for the cross-scale hybrid attention module described in section \ref{3.3}, enabling effective fusion and representation of contextual and fine-grained information for small object detection.

\subsection{Cross-Scale Hybrid Attention Module}\label{3.3}

In small object detection, feature sparsification caused by downsampling operations makes single-scale representations insufficient to provide adequate discriminative information. Moreover, conventional feature concatenation ignores the nonlinear spatial correspondences across feature maps of different scales.

\begin{figure}[htbp]
	\centering
	\includegraphics[width=\columnwidth]{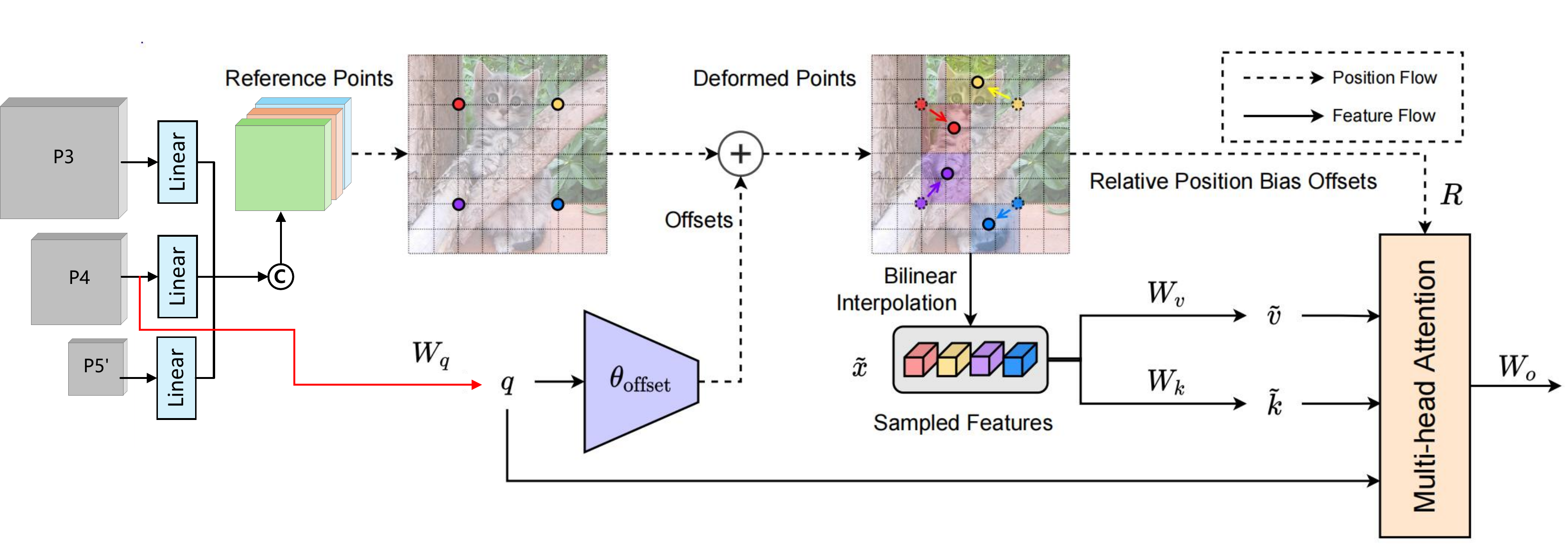}
	\caption{Cross-scale hybrid attention module (CSHA).}
	\label{fig3}
\end{figure}

To enable efficient multi-scale feature fusion and enhance sensitivity to fine-grained targets, we design and introduce a cross-scale hybrid attention module (CSHA). The proposed module takes features from the P4 level as queries and dynamically searches for relevant sampling locations across P3, P4, and P5 features enhanced by the global relation modeling module. By constructing a dynamic multi-scale feature fusion field, the module simultaneously captures high-resolution spatial details from P3 and high-level semantic information from P5. Unlike the global self-attention mechanism used in conventional Transformers, CSHA performs sparse sampling by learning offsets of key sampling points, which significantly reduces computational complexity while achieving effective cross-scale feature alignment, as illustrated in Fig.~\ref{fig3}.

\begin{figure}[htbp]
	\centering
	\includegraphics[width=\columnwidth]{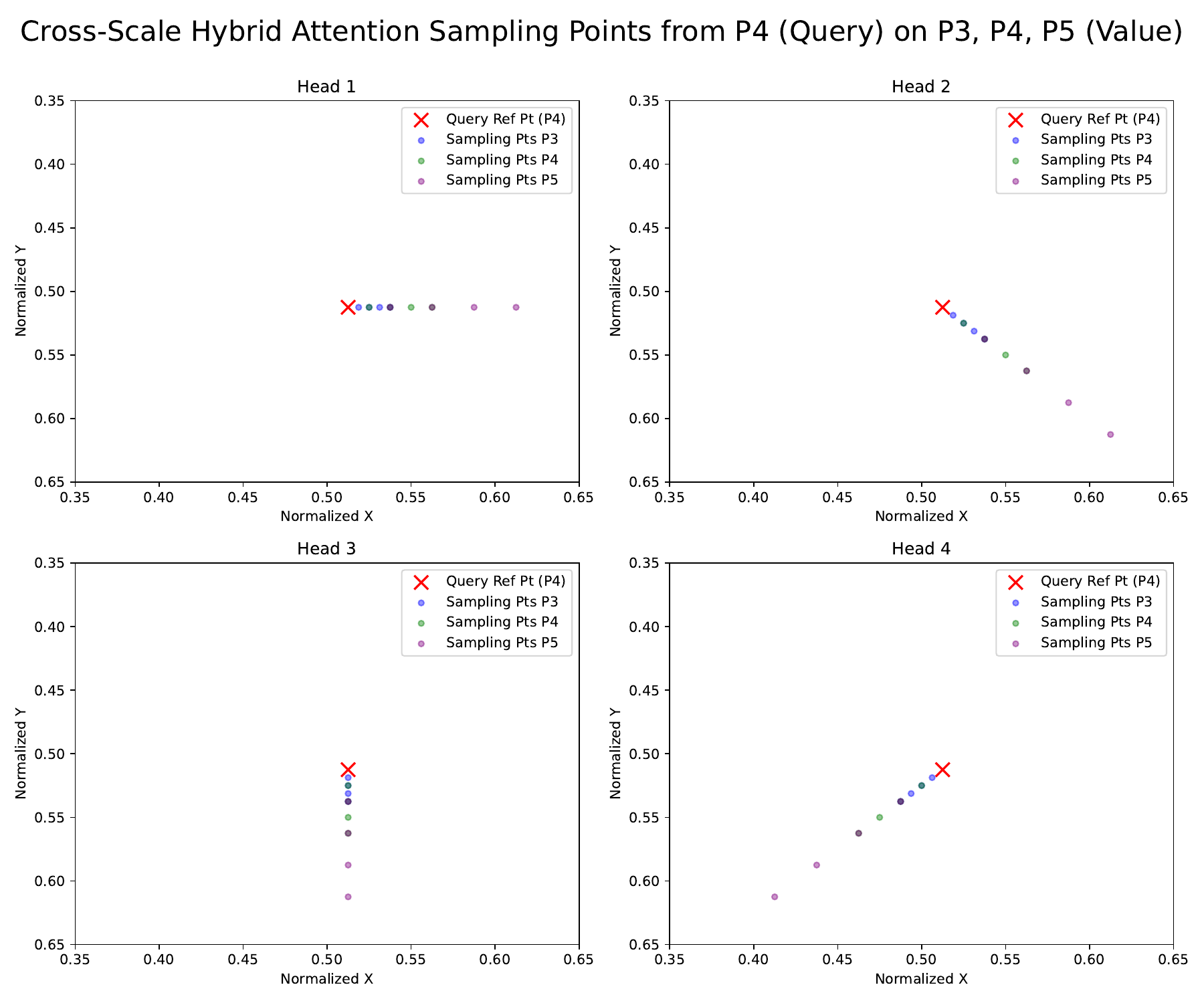}
	\caption{Schematic diagram of sampling points in P3, P4, and P5 Layers (Only 4 Heads of 8 are Shown).}
	\label{fig4}
\end{figure}

Specifically, features from different pyramid levels with channel dimensions $C_l$ are first projected into a unified embedding space of dimension $d$ using a set of $1 \times 1$ convolutional layers, ensuring that features interact within the same subspace. For each query pixel $q$ in the P4 feature map, a corresponding reference point $\hat{p}_q$ is determined in a normalized coordinate system, which are illustrated in Fig.~\ref{fig4}. Subsequently, an offset prediction network conditioned on the query feature and positional encoding predicts bilinear sampling offsets $\Delta p_{m l k}$ for each sampling point across three feature levels:
\begin{equation}
\Delta {p_{mlk}} = {W_{offset}} \cdot {{\mathbf{Z}}_q}
\end{equation}
where $m$ denotes the attention head, $l$ indicates the feature level, and $k$ indexes the sampling point. The final output feature is obtained by aggregating weighted sampling results from all feature levels:
\begin{equation}
	\begin{gathered}
		{\text{CSH - Attn}}(q,{{\hat p}_q},\{ {x^l}\} _{l = 3}^5) =  \hfill \\
		\sum\limits_{m = 1}^M {{W_m}} \left[ {\sum\limits_{l = 1}^L {\sum\limits_{k = 1}^K {{A_{mlk}}} }  \cdot {\phi _l}({x^l};{{\hat p}_q} + \Delta {p_{mlk}})} \right] \hfill \\ 
	\end{gathered}
\end{equation}
where $A_{m l k}$ represents the attention weight learned from the query, and ${\phi _l}\left(  \cdot  \right)$ denotes bilinear interpolation sampling performed on the corresponding feature level. The output resolution and channel dimensionality remain unchanged, enabling seamless integration with the original detection framework.

\subsection{Center-Assisted Loss Function}

Due to the high sensitivity of small objects to localization errors, IoU-based detectors often exhibit significant performance degradation when applied to small object detection. To address this limitation, we incorporate a center-assisted loss function into the bounding box regression branch, which accelerates network optimization by explicitly constraining the predicted bounding box centers \cite{tao2024edadet}. The center-assisted loss is defined as:
\begin{equation}
	\begin{gathered}
	D = \sqrt {{{({x_A} - {x_B})}^2} + {{({y_A} - {y_B})}^2}}  \hfill \\
	{L_{center\_assisted}} = 1 - {e^{ - D/C}} \hfill \\ 
\end{gathered}
\end{equation}
where $C$ represents the average object size in the dataset. The exponential normalization ensures that ${L_{center\_assisted}}$ is bounded within the range [0,1]. The final regression loss is formulated as a weighted combination of the center-assisted loss and the IoU loss:
\begin{equation}
	{L_{reg}} = {\alpha _1}{L_{center\_assisted}} + {\alpha _2}{L_{IoU}}
\end{equation}
where ${\alpha _1}$ and ${\alpha _2}$ are weighting coefficients that balance the contributions of the two loss terms. By providing effective gradient supervision even when IoU loss fails, the proposed center-assisted loss enhances training stability and improves localization accuracy for small objects.

\section{Experimental Results and Analysis}

\subsection{Dataset}

All experiments are conducted on the large-scale RGBT-Tiny benchmark dataset. RGBT-Tiny consists of 115 image sequences, among which 85 sequences are used for training and the remaining 30 sequences for testing. The dataset contains approximately 1.2 million manually annotated paired bounding box labels. All images are captured by onboard camera mounted on UAV, with a resolution of 512×640. Notably, more than 81\% of the objects are smaller than 16×16 pixels, making RGBT-Tiny particularly suitable for small object detection. The dataset covers seven object categories under diverse illumination conditions, with nighttime sequences accounting for 33.9\% of the total data. As a rigorous and comprehensive benchmark, RGBT-Tiny is adopted to evaluate the effectiveness and generalization capability of the proposed method.

\subsection{Experimental Settings}

All experiments are conducted on a server running Ubuntu 24.04, equipped with an NVIDIA GeForce RTX 5090 GPU. The implementation is based on Python 3.10 and PyTorch 2.1.0. During training, the stochastic gradient descent (SGD) optimizer is employed, and all models are trained for 100 epochs without using any pretrained weights. The initial learning rate is set to 0.01, with a weight decay of 5e-4 and a batch size of 16. All input images are resized to 640×640.

\subsection{Evaluation Metrics}

To comprehensively evaluate detection performance, we adopt evaluation metrics following the standard COCO evaluation protocol, including Average Precision (AP) metrics such as AP50 and mAP, which are defined as follows:
\begin{equation}
\left\{ \begin{gathered}
	AP = \int_0^1 {P(r)dr}  \hfill \\
	mAP = \frac{1}{N}\sum\nolimits_{i = 1}^N {A{P_i}}  \hfill \\ 
\end{gathered}  \right.
\end{equation}

In addition, we adopt the Scale-Adaptive Fitness (SAFit) metric introduced in \cite{3}, which exhibits strong robustness for both small and large objects. SAFit is formulated as a weighted combination of IoU and the Normalized Wasserstein Distance (NWD):
\begin{equation}
\begin{gathered}
	{\text{SAFit}} =  \hfill \\
	\frac{1}{{1 + {e^{ - (\sqrt A /C - 1)}}}} \cdot {\text{IoU}}{\kern 1pt}  + \left( {1 - \frac{1}{{1 + {e^{ - (\sqrt A /C - 1)}}}}} \right) \cdot {\text{NWD}} \hfill \\ 
\end{gathered}
\end{equation}

\subsection{Comparison with the State-of-the-Art}

Tables \ref{tab_5} and \ref{tab_6} report quantitative comparisons with recent state-of-the-art detectors on the RGBT-Tiny dataset under IoU-based and SAFit-based evaluation metrics, respectively.

Under the standard IoU-based COCO metrics, our method achieves the highest performance, obtaining an AP of 21.4, AP50 of 45.4, and AP75 of 18.1, outperforming all competing approaches, including both anchor-based and transformer-based detectors. Notably, our approach surpasses diffusion-based and DETR-style models while maintaining a moderate model size.
\begin{table}[h]
	\footnotesize
	\centering
	\caption{IoU-based results comparison with other advanced methods on RGBT-Tiny using COCO metrics. The top two results are presented in bold and underlined format.}\label{tab_5}
	\renewcommand\arraystretch{1.2}
	\begin{tabular*}{\linewidth}{@{\extracolsep{\fill}}l|c|c|ccc}
		\toprule[1.5pt]
		Methods&Year&Param.&AP&AP$_{50}$&AP$_{75}$\\
		\midrule[1pt]
		YOLO \cite{YOLO}&2016&61.5M&10.0 &26.7 &5.1\\
		Cascade RCNN \cite{CascadeRCNN}&2018&68.9M&{15.0} &{33.9} &{11.1}\\
		Dynamic RCNN \cite{DynamicRCNN}&2020&41.2M&14.5 &33.5 &10.7\\
		Deformable DETR \cite{Deformable-DETR}&2020&39.8M&12.3 &30.9 &7.3\\
		TOOD \cite{TOOD}&2021&\textbf{31.8M}&13.0 &30.3 &9.9\\
		VarifocalNet \cite{VarifocalNet}&2021&\underline{32.5M}&13.0 &29.4 &10.1\\
		{DINO \cite{DINO}}&2022&97.7M&17.0&39.6&11.7\\
		Sparse RCNN \cite{SparseRCNN}&2023&44.2M&9.1 &21.4 &6.4\\
		{CO-DETR\cite{CO-DETR}}&2023&65.2M&17.8&39.1&14.6\\
		{DiffusionDet \cite{DiffusionDet}}&2023&151.0M&\underline{19.7}&\underline{42.4}&\underline{16.2}\\
		{DDQ \cite{DDQ}}&2023&48.3M&15.2&34.6&11.7\\
		CMA-Det \cite{CMA-Det}&2024&33.4M&5.5&16.1&2.2 \\
		\midrule[0.5pt]
		\multicolumn{2}{l|}{Ours}&58.2M&\textbf{21.4}&\textbf{45.4}&\textbf{18.1} \\
		\bottomrule[1.5pt]
	\end{tabular*}
\end{table}

\begin{table}[h]
	\footnotesize
	\centering
	\caption{SAFit-based results comparison with other advanced methods on RGBT-Tiny using COCO metrics. The top two results are presented in bold and underlined format.}\label{tab_6}
	\renewcommand\arraystretch{1.2} 
	\begin{tabular*}{\linewidth}{@{\extracolsep{\fill}}l|c|c|ccc}
		\toprule[1.5pt]
		Methods&Year&Param.&AP&AP$_{50}$&AP$_{75}$\\
		\midrule[1pt]
		YOLO \cite{YOLO}&2016&61.5M&24.3 &37.7 &28.4 \\
		Cascade RCNN \cite{CascadeRCNN}&2018&68.9M&30.1 &44.2 &35.8 \\
		Dynamic RCNN \cite{DynamicRCNN}&2020&41.2M&29.4 &44.0 &34.2 \\
		Deformable DETR \cite{Deformable-DETR}&2020&39.8M&28.2 &{45.4} &32.0 \\
		TOOD \cite{TOOD}&2021&\textbf{31.8M}&27.9 &43.5 &31.7 \\
		VarifocalNet \cite{VarifocalNet}&2021&\underline{32.5M}&26.9 &41.6 &30.1 \\
		{DINO \cite{DINO}}&2022&{97.7M}&{34.7}&52.4&41.3 \\
		Sparse RCNN \cite{SparseRCNN}&2023&44.2M&19.2 &29.8 &21.9 \\
		{CO-DETR\cite{CO-DETR}}&2023&{65.2M}&35.0&{52.0}&{40.9} \\
		{DiffusionDet \cite{DiffusionDet}}&2023&{151.0M}&\underline{38.4}&\underline{55.7}&\underline{45.5} \\
		{DDQ \cite{DDQ}}&2023&{48.3M}&{31.5}&{47.8}&{37.0} \\
		CMA-Det \cite{CMA-Det}&2024&33.4M&18.9&40.9&16.4 \\
		\midrule[0.5pt]
		\multicolumn{2}{l|}{Ours}&58.2M&\textbf{40.1}&\textbf{57.7}&\textbf{47.7} \\
		\bottomrule[1.5pt]
	\end{tabular*}
\end{table}

\begin{figure}[h]
	\centering
	\includegraphics[width=\columnwidth]{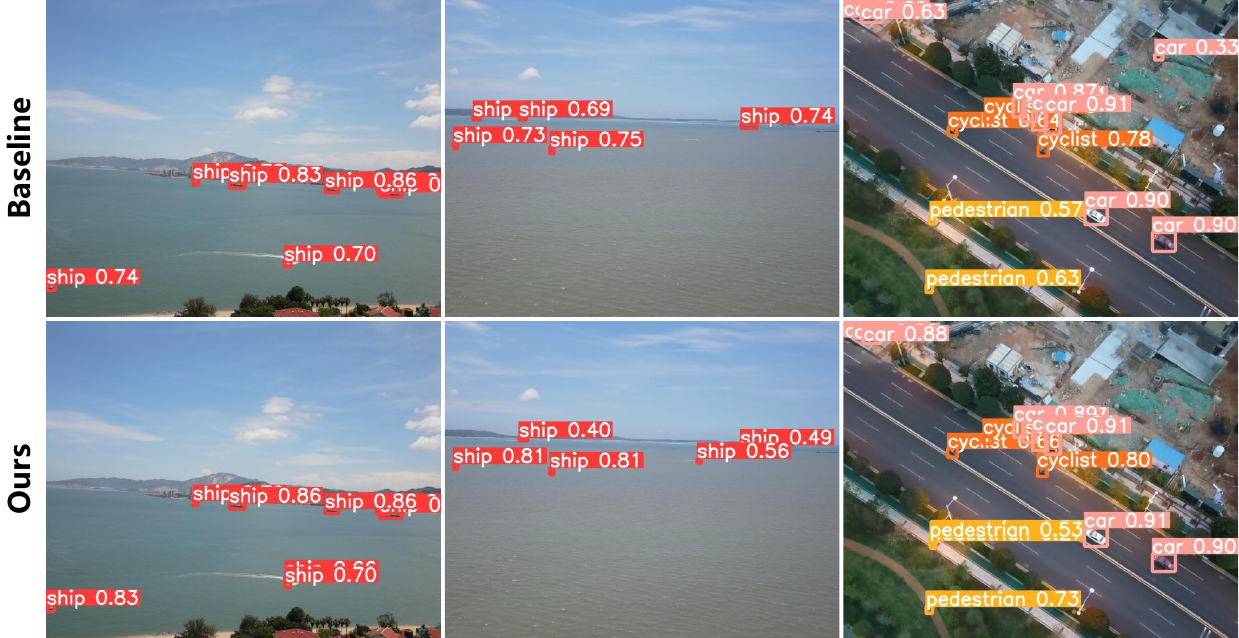}
	\caption{Visualized comparison of results between the baseline and our method on the RGBT-Tiny dataset.}
	\label{fig5}
\end{figure}

When evaluated using the SAFit-based metrics, which is more robust for small object localization, our method consistently achieves the best results across all metrics, reaching an AP of 40.1, AP50 of 57.7, and AP75 of 47.7. These improvements demonstrate the effectiveness of the proposed framework in accurately localizing small objects and highlight its strong generalization capability under scale-sensitive evaluation criteria. Overall, the results confirm that our method establishes a new state-of-the-art performance on the RGBT-Tiny benchmark. Fig.~\ref{fig5} presents a visualized comparison of the results obtained by the baseline and our method on the RGBT-Tiny dataset.

\subsection{Ablation Study}

To evaluate the effectiveness of each proposed component, we conduct a comprehensive ablation study on the RGBT-Tiny dataset. The results are summarized in Table \ref{tab_1}, where the baseline model (Yolov5-l) is progressively enhanced by incorporating different modules.
\begin{table}[h]
	\footnotesize
	\centering
	\caption{IoU-based ablation study results on RGBT-Tiny.}\label{tab_1}
	\renewcommand\arraystretch{1.2}
	\begin{tabular*}{\linewidth}{@{\extracolsep{\fill}}l|cc|cc}
		\toprule[1.5pt]
		Methods&Param.&GFLOPs&AP$_{50}$&AP\\
		\midrule[1pt]
		Baseline&46.1M&107.7&46.1&21.6\\
		+ RHWD Module&46.1M&108.1&47.4&21.7\\
		+ GRM Module&55.6M&113.1&47.9&22.1\\
		+ CSHA Module&58.2M&125.1&48.2&22.9\\
		+ Center-Assisted Loss&58.2M&125.1&\textbf{48.4}&\textbf{23.0}\\
		\bottomrule[1.5pt]
	\end{tabular*}
\end{table}

Starting from the baseline detector, introducing the RHWD module yields a noticeable improvement, increasing AP50 from 46.1 to 47.4 with only a marginal increase in computational cost. This gain indicates that integrating frequency-domain information with spatial features effectively alleviates the loss of fine-grained details during downsampling, which is particularly beneficial for small object representation. By further incorporating the GRM module, AP50 improves to 47.9, demonstrating the importance of long-range dependency modeling and global semantic aggregation in suppressing background interference. Although this module introduces additional parameters, the performance gain justifies the increased model capacity. The addition of the CSHA module leads to a more significant improvement, boosting AP50 to 48.2 and AP to 22.9. This result confirms that dynamically aligned cross-scale feature interactions play a critical role in effectively fusing high-resolution spatial details and high-level semantic information, thereby enhancing small object detection performance. Finally, incorporating the Center-Assisted Loss further improves localization accuracy, achieving the best overall performance with an AP50 of 48.4 and an AP of 23.0, without introducing any additional computational overhead during inference. This demonstrates that the proposed loss function provides effective supervision for small object regression, particularly in scenarios where IoU-based losses alone are insufficient. Overall, the ablation results validate that each component contributes positively to the detection performance, and their combination yields a cumulative and complementary effect, leading to consistent improvements over the baseline.
\begin{table}
	\footnotesize
	\centering
	\caption{IoU-based results comparison of the RHWD Module with Large-kernel Convolution and Focus on RGBT-Tiny.}\label{tab_2}
	\renewcommand\arraystretch{1.2}
	\begin{tabular*}{\linewidth}{@{\extracolsep{\fill}}l|cc|cc}
		\toprule[1.5pt]
		Methods&Param.&GFLOPs&AP$_{50}$&AP\\
		\midrule[1pt]
		Large-kernel Convolution&46.1M&107.7&46.1&21.6\\
		Focus&46.1M&107.7&41.7&19.5\\
		RHWD Module&46.1M&108.1&\textbf{47.4}&\textbf{21.7}\\
		\bottomrule[1.5pt]
	\end{tabular*}
\end{table}

\noindent{\textbf{Effect of RHWD Module.}} Table \ref{tab_2} compares the proposed RHWD module with large-kernel convolution and the Focus operation on the RGBT-Tiny dataset. Under identical parameter and computational budgets, the RHWD module achieves the best performance, improving AP from 46.1 to 47.4. RHWD more effectively retains fine-grained structural details by jointly leveraging spatial and frequency-domain representations, demonstrating its superiority for small object feature extraction.

\noindent{\textbf{Effect of GRM Module.}} Table \ref{tab_3} presents a comparison between the proposed GRM module and conventional multi-head self-attention on the RGBT-Tiny dataset. While directly applying multi-head self-attention increases model complexity, it leads to a degradation in AP, indicating that normal global attention is insufficient for robust small object detection. In contrast, the GRM module achieves the best performance, improving AP50 to 47.9 and AP to 22.1, demonstrating its effectiveness in aggregating global contextual information while suppressing background interference.
\begin{table}
	\footnotesize
	\centering
	\caption{IoU-based results comparison of the GRM Module with Multi-Head Self-Attention on RGBT-Tiny.}\label{tab_3}
	\renewcommand\arraystretch{1.2}
	\begin{tabular*}{\linewidth}{@{\extracolsep{\fill}}l|cc|cc}
		\toprule[1.5pt]
		Methods&Param.&GFLOPs&AP$_{50}$&AP\\
		\midrule[1pt]
		without&46.1M&108.1&47.4&21.7\\
		Multi-Head Self-Attention&52.4M&112.3&46.3&21.9\\
		GRM Module&55.6M&113.1&\textbf{47.9}&\textbf{22.1}\\
		\bottomrule[1.5pt]
	\end{tabular*}
\end{table}

\begin{table}
	\footnotesize
	\centering
	\caption{IoU-based results comparison of the GRM Module placed Before and After SPPF on RGBT-Tiny.}\label{tab_4}
	\renewcommand\arraystretch{1.2}
	\begin{tabular*}{\linewidth}{@{\extracolsep{\fill}}l|cc|cc}
		\toprule[1.5pt]
		Methods&Param.&GFLOPs&AP$_{50}$&AP\\
		\midrule[1pt]
		without&46.1M&108.1&47.4&21.7\\
		Before SPPF Module&55.6M&113.1&46.4&21.6\\
		After SPPF Module&55.6M&113.1&\textbf{47.9}&\textbf{22.1}\\
		\bottomrule[1.5pt]
	\end{tabular*}
\end{table}

\noindent{\textbf{Effect of GRM Module's Position.}} Table \ref{tab_4} investigates the impact of inserting the Global Relation Modeling (GRM) module at different positions in the backbone. When the GRM module is placed before the SPPF layer, the detection performance slightly degrades compared with the baseline, indicating that global relation modeling on insufficiently aggregated features may introduce noise. In contrast, placing it after the SPPF layer yields the best results, improving AP50 to 47.9 and AP to 22.1. This confirms that applying global relation modeling on high-level, semantically rich features is more effective for enhancing small object detection.

\section{Conclusion}
In this paper, we proposed a multi-level feature enhancement and global relation modeling framework tailored for small object detection under complex backgrounds. By integrating residual wavelet-based downsampling, global semantic relation modeling, cross-scale hybrid attention, and a center-assisted regression loss, the proposed method effectively enhances feature representation, multi-scale alignment, and localization stability for small objects. Extensive experiments on the RGBT-Tiny benchmark demonstrate that our approach consistently outperforms existing state-of-the-art methods under both IoU-based and scale-adaptive evaluation metrics. Future work will explore extending the proposed framework to broader multi-modal detection scenarios and further improving efficiency for real-time applications.


\bibliographystyle{IEEEtran}
\bibliography{Reference.bib}

\end{document}